\documentclass{article}

\usepackage[numbers,sort&compress]{natbib}

\usepackage[final]{neurips_2024}

\usepackage[utf8]{inputenc} 
\usepackage[T1]{fontenc}    
\usepackage{hyperref}       
\usepackage{url}            
\usepackage{booktabs}       
\usepackage{amsfonts}       
\usepackage{nicefrac}       
\usepackage{microtype}      
\usepackage{xcolor}         
\usepackage{graphicx}
\usepackage{amsmath}
\usepackage{amssymb}
\usepackage{dsfont}

\usepackage{wrapfig}
\usepackage{algorithm}
\usepackage{algpseudocode}
\usepackage{multirow}

\usepackage[capitalize]{cleveref}
\crefname{section}{Sec.}{Secs.}
\Crefname{section}{Section}{Sections}
\Crefname{table}{Table}{Tables}
\crefname{table}{Tab.}{Tabs.}

\usepackage{pifont} 

\newcommand{\cmark}{\ding{51}}%
\newcommand{\xmark}{\ding{55}}%

\title{Rethinking The Training And Evaluation of Rich-Context Layout-to-Image Generation}

\author{
Jiaxin Cheng$^2$\thanks{Work done during his internship at Amazon} \quad Zixu Zhao$^1$ \quad Tong He$^1$ \quad Tianjun Xiao$^1$  \quad Zheng Zhang$^1$ \quad Yicong Zhou$^2$\\
\texttt{\{yc47434,yicongzhou\}@um.edu.mo} \texttt{\{zhaozixu,tianjux,htong,zhaz\}@amazon.com}\\
$^1$Amazon Web Services Shanghai AI Lab \qquad $^2$University of Macau 
}

\begin{document}

\maketitle

\begin{figure}[!h]
    \centering
    \includegraphics[width=0.95\linewidth]{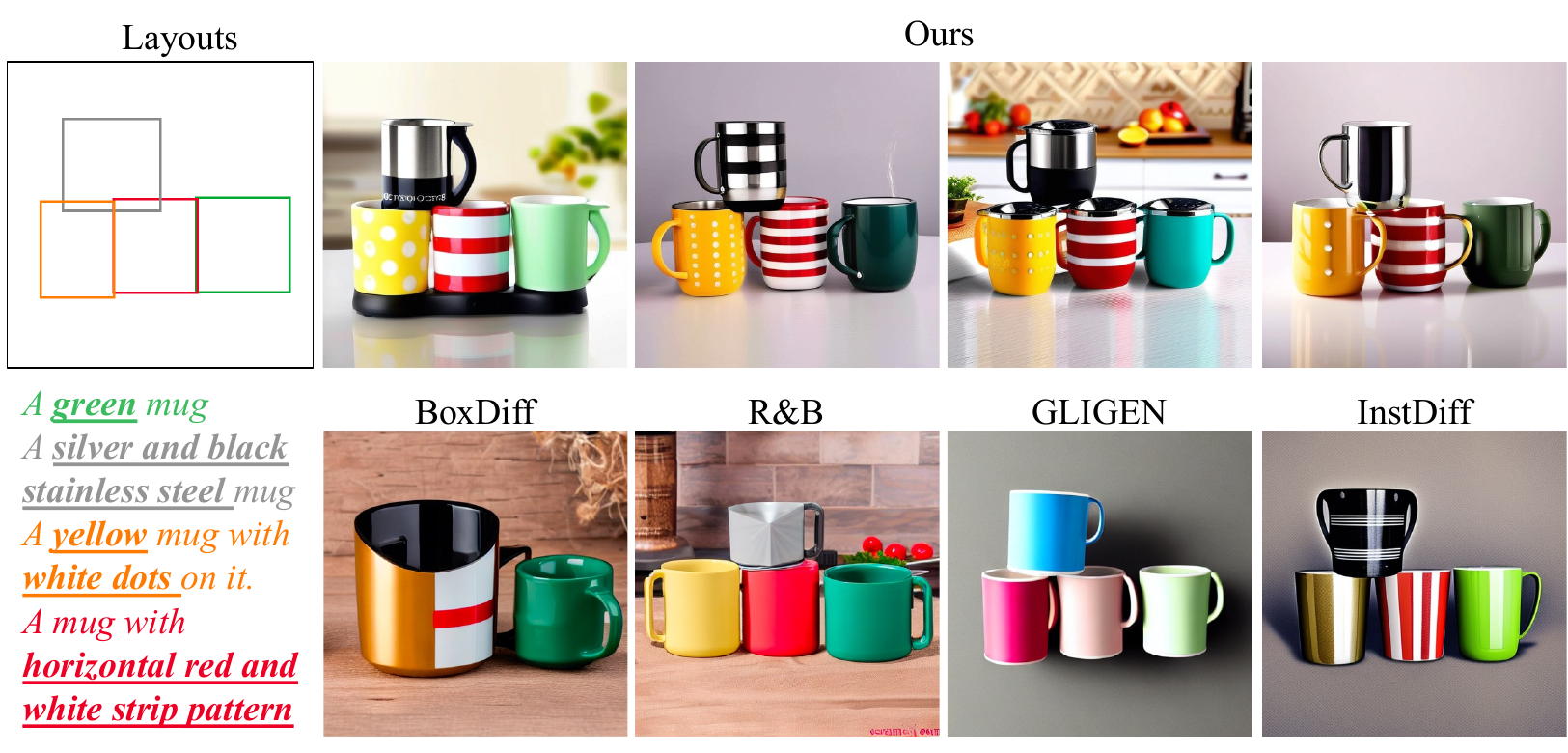}
    \caption{The proposed method demonstrates the ability to accurately generate objects with complex descriptions in the correct locations while faithfully preserving the details specified in the text. In contrast, existing methods such as BoxDiff~\cite{boxdiff}, R\&B~\cite{randb}, GLIGEN~\cite{li2023gligen}, and InstDiff~\cite{instdiff} struggle with the complex object descriptions, leading to errors in the generated objects. }
    \label{fig.teaser}
\end{figure}

\begin{abstract}

Recent advancements in generative models have significantly enhanced their capacity for image generation, enabling a wide range of applications such as image editing, completion and video editing. A specialized area within generative modeling is layout-to-image (L2I) generation, where predefined layouts of objects guide the generative process. In this study, we introduce a novel regional cross-attention module tailored to enrich layout-to-image generation. This module notably improves the representation of layout regions, particularly in scenarios where existing methods struggle with highly complex and detailed textual descriptions. Moreover, while current open-vocabulary L2I methods are trained in an open-set setting, their evaluations often occur in closed-set environments. To bridge this gap, we propose two metrics to assess L2I performance in open-vocabulary scenarios. Additionally, we conduct a comprehensive user study to validate the consistency of these metrics with human preferences.  \url{https://github.com/cplusx/rich_context_L2I/tree/main}
\end{abstract}

\section{Introduction}


Recent years have witnessed significant advancements in image generation, with the diffusion model~\cite{sohl2015deep, ddpm} emerging as a leading method. This model has shown scalability with billion-scale web training data and has achieved remarkable quality in text-to-image generation tasks~\cite{ldm,dalle2,dalle3,glide,podell2023sdxl}. However, text-to-image models that rely solely on textual descriptions face limitations, particularly in scenarios requiring precise location control. 

As the diversity and complexity of model training tasks increase, there is a growing demand for both accuracy and precision in generated data. Precision involves more accurate object positioning, while accuracy ensures that generated objects closely match intricate descriptions, even in highly complex scenarios. Recent approaches~\cite{li2023gligen,instdiff} have addressed this by incorporating precise location control into the diffusion model, enabling open-vocabulary layout-to-image (L2I) generation. Among various layout types, bounding box based layouts offer intuitive and convenient control compared to masks or keypoints~\cite{ldm}. Additionally, bounding box layouts provide greater flexibility for diverse and detailed descriptions. In this work, we systematically investigate layout-to-image generation from bounding box-based layouts with rich context, where the descriptions for each instance to be generated can be complex, lengthy, and diverse, aiming to produce highly accurate objects with intricate and detailed descriptions.


Revisiting existing diffusion-based layout-to-image generation methods reveals that many rely on an extended self-attention mechanism~\cite{li2023gligen,instdiff}, which applies self-attention to the combined features of visual and textual tokens. This approach condenses textual descriptions for individual objects into single vectors and aligns them with image features through a dense connected layer.

However, a closer examination of how diffusion models achieve text-to-image generation~\cite{ldm,glide,podell2023sdxl,dalle2,pixart_alpha} shows that text conditions are typically integrated via cross-attention layers rather than self-attention layers. Adopting cross-attention preserves text features as sequences of token embeddings instead of consolidating them into a single vector. Recent diffusion models have demonstrated improved generation results by utilizing larger~\cite{pixart_alpha,pixart_sigma} or multiple~\cite{podell2023sdxl} text encoders and more detailed image captions~\cite{dalle3}. This underscores the significance of the cross-attention mechanism in enhancing generation quality through richer text representations and more comprehensive textual information.


Drawing an analogy between the generation of individual objects and the entire image, it is natural to consider applying similar cross-attention mechanisms to each object. Therefore, we propose introducing Regional Cross-Attention modules for layout-to-image generation, enabling each object to undergo a generation process akin to that of the entire image.


In addition to the proposed training scheme for L2I generation, we have identified a lack of reliable evaluation metrics for open-vocabulary L2I generation. While models~\cite{li2023gligen,instdiff} can perform open-vocabulary L2I generation, evaluations are typically conducted on closed-set datasets such as COCO~\cite{coco} or LVIS~\cite{gupta2019lvis}. However, such closed-set evaluations may not accurately reflect the capabilities of open-vocabulary L2I models, as the text descriptions in these datasets are often limited to just a few words. It remains unclear whether these models can perform effectively when presented with complex and detailed object descriptions.

To address this gap, we propose two metrics that consider object-text alignment and layout fidelity that works for rich-context descriptions. Additionally, we conduct a user study to assess the reliability of these metrics and identify the circumstances under which these metrics may fail to reflect human preferences accurately.

Our contributions can be summarized as follows: 1) We revisit the training of L2I generative models and propose regional cross-attention module to enhance rich-context L2I generation, outperforming existing self-attention-based approaches.
2) To effectively evaluate the performance of open-set L2I models, we introduce two metrics that assess the models' capabilities with rich-context object descriptions and validate their reliability through a user study.
3) Our experimental results demonstrate that our proposed solution improves generation performance, especially with rich-context prompts, while reducing computational cost in each layout-conditioning layer thanks to the use of cross-attention.

\section{Related Works}

\textbf{Diffusion-based Generative Models} The emergence of diffusion models~\cite{sohl2015deep, ddpm} has significantly advanced the field of image generation. Within just a few years, diffusion models have made remarkable progress across various domains, including super-resolution~\cite{saharia2022image}, colorization~\cite{saharia2022palette}, novel view synthesis~\cite{watson2022novel}, 3D generation~\cite{poole2022dreamfusion, tang2023make, cheng2023sdfusion}, image editing~\cite{meng2021sdedit,brooks2023instructpix2pix,kawar2023imagic}, image completion~\cite{palette} and video editing~\cite{singer2022make,cheng2023consistent}. This progress can be attributed to several factors. Enhancements in network architectures~\cite{ldm,glide,dalle2,podell2023sdxl,imagen,pernias2024wrstchen} have played a pivotal role. Additionally, improvements in training paradigms~\cite{nichol2021improved, song2019generative, dhariwal2021diffusion, song2020score} have contributed to this advancement. Moreover, the ability to incorporate various conditions during image generation has broadened the impact and applications of diffusion models. These conditions include elements such as segmentation~\cite{avrahami2023spatext, avrahami2022blended, balaji2022ediffi, yang2023paint}, using an image as a reference~\cite{mou2023t2i, ruiz2023dreambooth, textualinversion}, and layout~\cite{cheng2023layoutdiffuse, li2023gligen, instdiff}, the latter of which will be the main focus of our discussion in this work.

\noindent\textbf{Layout-to-image generation}: Early works~\cite{lostgan,ocgan,lama,context_l2i,spade,pSp,taming,twfa} often utilized GANs~\cite{gan} or transformers~\cite{allyouneed} for L2I generation. For instance, GAN-based LAMA~\cite{lama}, LostGANs~\cite{lostgan}, and Context L2I~\cite{context_l2i} encode layouts as style features fed into adaptive normalization layers, while Taming~\cite{taming} and TwFA~\cite{twfa} use transformers to predict latent visual codes from pretrained VQ-VAE~\cite{vqvae}. Recent diffusion models~\cite{cheng2023layoutdiffuse,li2023gligen,instdiff,boxdiff,randb,xue2023freestyle,zheng2023layoutdiffusion} have shown promising results, extending L2I generation to be open-set. LayoutDiffuse~\cite{cheng2023layoutdiffuse}  injects objects into the image features through learning per-class embeddings. LayoutDiffusion~\cite{zheng2023layoutdiffusion} fine-tunes pre-trained diffusion models by mapping object labels and layout coordinates into cross-attendable embeddings for attention layers. FreestyleL2I~\cite{xue2023freestyle}, BoxDiff~\cite{boxdiff} and R\&B~\cite{randb}, which are training-free methods, leverage pre-trained diffusion models to inject objects into specified regions by imposing spatial constraints. GLIGEN~\cite{li2023gligen} and InstDiff~\cite{instdiff} explore open-set L2I generation using grounded bounding boxes, which encodes layout locations and object descriptions into features attended by self-attention layers.

\section{Methodology}

\subsection{Challenges in Rich-Context Layout-to-Image Generation }

\begin{figure}[t]
    \centering
    \includegraphics[width=0.95\linewidth]{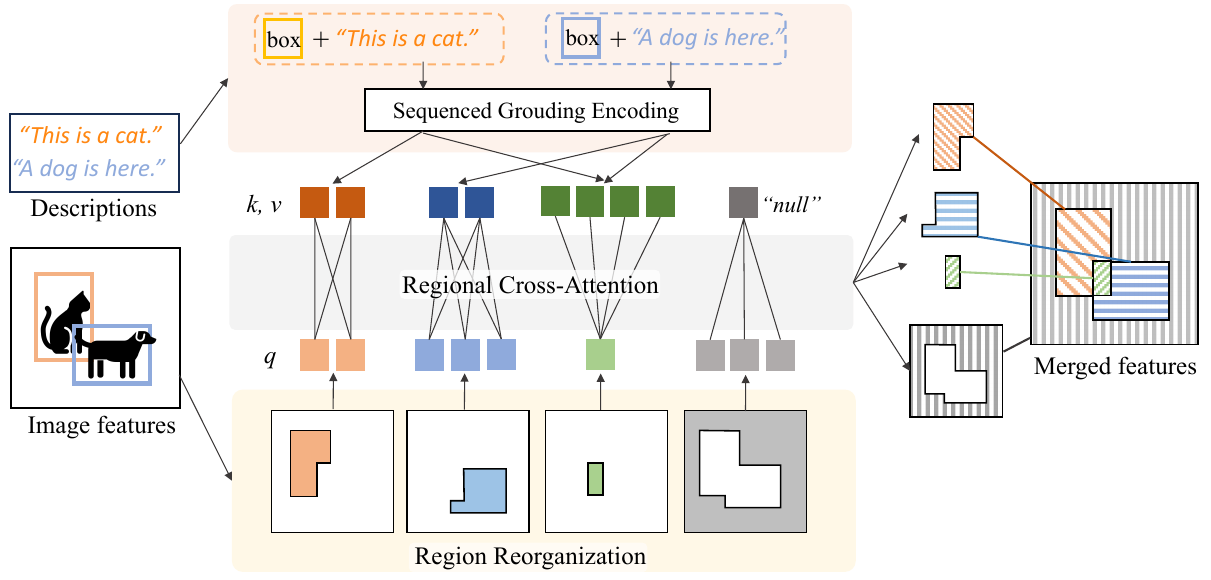}
    \caption{An example of regional cross-attention with two overlapping objects. Cross-attention is applied to each pair of regional visual and grounded textual tokens.  The overlapping region cross-attends with the textual tokens containing both objects, while the non-object region attends to a learnable ``$null$'' token.}
    \label{fig.regional_xattn}
\end{figure}

The layout-to-image (L2I) generation task can be formally defined as follows: given a set of description tuples $S := \{s_i | s_i=(b_i, t_i)\}$, where $b_i$ represents the bounding box coordinates of an object, and $t_i$ denotes the corresponding text description,  the objective is to generate an image that accurately aligns objects with their respective descriptions while maintaining fidelity to the specified layouts. In the closed-set setting, the number of text descriptions is limited to a fixed number $N$, \textit{i.e.}, $N = |\{t_i\}|$, where $N$ is the total number of classes. However, in the open-set and rich-context settings, the number of descriptions is unlimited, with descriptions in the rich-context setting being more diverse, complex, and lengthy.

Rich-context L2I encounters several challenges: 
1) The rich-context descriptions for each object can be lengthy and complex, requiring the model to correctly understand the descriptions without overlooking details. Existing open-set layout-to-image solutions~\cite{li2023gligen,instdiff} typically condense and map text embeddings into a single vector, which is then mapped to the image space for layout conditioning. However, this condensation process can result in significant information loss, particularly for lengthy descriptions. 2) Fitting various text descriptions into their designated layout boxes while maintaining global consistency is challenging. Unlike simpler text-to-image generation with a single description, L2I generation deals with multiple objects, requiring precise matching of each description to its specific layout area without causing global inconsistency. 3) L2I involves objects with intersecting bounding boxes, unlike segmentation-mask-to-image tasks where object areas do not overlap and can be efficiently handled by pixel-conditioned methods such as ControlNet~\cite{controlnet} and Palette~\cite{palette}. L2I models must determine the appropriate order and occlusion of overlapping objects autonomously, ensuring the proper representation and interaction of each object within the image.

\subsection{Regional Cross-Attention}

We propose using a regional cross-attention layer as a solution to rich-context layout-to-image generation, addressing the aforementioned challenges. The desired properties for an effective rich-context layout-conditioning module are as follows: 1) \textit{Flexibility}: The model must accurately understand rich-context descriptions, regardless of their length or complexity, ensuring that no details are overlooked. 2) \textit{Locality}: Each textual token should only attend to the visual tokens within its corresponding layout region, without influencing regions beyond the layout. 3) \textit{Completeness}: All visual features, including those in the background, should be properly attended by certain description to maintain consistency in the output feature distribution. 
4) \textit{Collectiveness}: In cases where a visual token overlaps with multiple objects, it should consider all descriptions related to those intersecting objects.

Our approach differentiates itself from previous methods~\cite{li2023gligen,instdiff} by employing cross-attention layers, rather than self-attention layers, to condition objects within the image. This design is inspired by the architecture of modern text-to-image diffusion models, which achieve fine-grained text control by incorporating pre-pooled textual features in the cross-attention layers. Analogously, one can apply cross-attention repeatedly between pre-pooled object description tokens and visual tokens within the corresponding regions for all objects. However, this straightforward method, though satisfies flexibility and locality, does not fully meet the criteria of completeness and collectiveness, as it may inadequately address non-object regions and overlapping objects. This limitation can result in inconsistent global appearances and challenges in managing overlapping objects effectively.

\noindent\textbf{Region Reorganization.} We propose region reorganization to satisfy locality, completeness and collectiveness, by creating a spatial partition of the image based on the layout. Each region is classified into one of three types: single-object region, overlapping region among objects, and background. This partitioning ensures that regions are mutually exclusive (\textit{i.e.}, non-overlapping). Figure~\ref{fig.regional_xattn} illustrates a simple case with two overlapping objects. The overlapping area becomes a new, distinct region, while the non-overlapping parts of the original regions and remaining background are also treated as separate, new regions, thus ensuring completeness.

Formally, in the general case with multiple objects, the reorganized regions $R := \{r_i\}$ satisfy that the union of these regions will form a complete mask covering the entire visual feature space, while ensuring no intersection between any two reorganized regions:

\begin{align}
\bigcup_{i=1}^{|R|} r_i &= \mathds{1}; & r_i \cap r_j &= \varnothing \quad \text{for } i \neq j \text{ and } i, j \in [1, |R|]
\end{align}

Our regional cross-attention operates within each reorganized region. We define a selection operation $f(\cdot, r_i)$ to identify the appropriate regions for cross-attention. For visual tokens $V := \{v_j\}$ it finds the tokens whose locations $\mathrm{loc}(v_j)$ lie within the $i$-th reorganized region. For description tuples $S$, it filters the instances that overlap with the $i$-th reorganized region. This selection operation ensures that the text description is applied exclusively to the visual tokens within its corresponding region, thus maintaining locality. For regions with multiple objects, $f$ also ensures that all overlapping descriptions are included to satisfy collectiveness.

\begin{align}
    f(V, r_i) &:= \{v_j | \mathrm{loc}(v_j) \in r_i \} \\
    f(S, r_i) &:= \{s_j | b_j \cap r_i \neq \varnothing \}
\end{align}

\begin{wrapfigure}{r}{0.45\textwidth}
    \centering
    \includegraphics[width=0.45\textwidth]{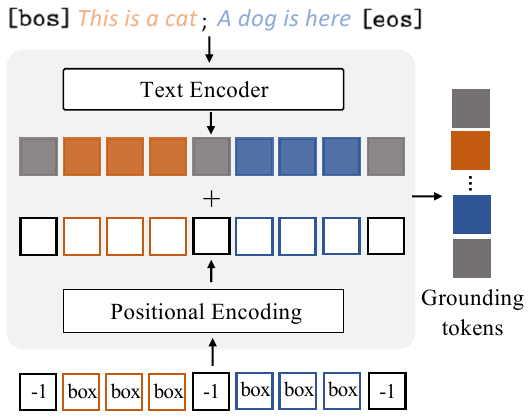}
    \caption{Sequenced Grounding Encoding with box coordinates as indicators.}
    \label{fig.text_seq}

\end{wrapfigure}

The final attention result $A$ is the aggregation of all regional attention outputs. The selected descriptions $f(S, r_i)$ are encoded using Sequenced Grounding Encoding (SGE) in \Cref{fig.text_seq} and serve as the key and value during cross-attention. For non-object regions where $f(S, r_i) = \varnothing$, a ``$null$'' embedding is learned as a substitute for the description.

\begin{align}
    a_i &= \mathrm{CrossAttn}(f(V, r_i), \mathrm{SGE}[f(S, r_i)]); &
    A &= \bigcup_{i=1}^{|R|} a_i
\end{align}

\noindent\textbf{Sequenced Grounding Encoding with Box Indicator.} The selected object descriptions in each reorganized region are encoded into textual tokens using Sequenced Grounding Encoding in \Cref{fig.text_seq}. When multiple objects are present in a region, their descriptions are concatenated with a separator token. However, if two objects share the same description, their encoded textual embeddings will be identical. This identical embedding makes it impossible for the cross-attention module to distinguish between distinct objects. To address this issue, we incorporate bounding box coordinates as additional indicators. During encoding, we concatenate the bounding box coordinates with the textual tokens. The bounding box coordinates are initially encoded using sinusoidal positional encoding~\cite{allyouneed} and then repeated to match the length of the textual tokens before concatenation. For separator tokens and special tokens such as \texttt{[bos]} and \texttt{[eos]}, we use an all -1 vector for their box coordinates.

\section{Evaluation for Rich-Context L2I Generation}


\subsection{Rethinking L2I Evaluation}


For evaluating layout-to-image generation, what are mainly considered are two aspects: 1) Object-label alignment, which checks whether the generated object matches the corresponding descriptions. 2) Layout fidelity, which examines how well the generated object aligns with the given bounding box.

In closed-set scenarios, it is common to use an off-the-shelf detector to evaluate L2I generation performance~\cite{cheng2023layoutdiffuse,li2023gligen,instdiff}. Object-label alignment is assessed by classifying image crops extracted from the generated image using a pre-trained classifier. Similarly, layout fidelity is measured by comparing the bounding boxes detected in the generated image with the provided layouts, using a pre-trained object detector.

However, in the open-set scenario, it is impossible to list all the classes. Moreover, even the state-of-the-art open-set object detectors~\cite{yoloworld,zareian2021open,yao2022detclip} are set up to handle inputs at the word or phrase level, which falls short for the sentence-level descriptions required in rich-context L2I generation. we introduce two metrics to bridge the gap in evaluating open-vocabulary L2I models.

\subsection{Metrics For Rich-Context L2I}\label{sec.metric}

We leverage the powerful visual-textual model CLIP for measuring object-label alignment, and the Segment Anything Model (SAM) for evaluating layout fidelity of the generated objects.

\textbf{Crop CLIP Similarity}: In rich-context L2I, object descriptions can be diverse and complex. 
The CLIP model, known for its robustness in image-text alignment, is thus suitable for this evaluation. To ensure accuracy and mitigate interference from surrounding objects, we compute the CLIP score after cropping the object as per the layout specifications. 


\textbf{SAMIoU}: An accurately generated object should closely align with its designated layout. Given the potential diversity in object shapes, we employ the SAM model, which can highlight an object's region in mask format within a given box region, to identify the actual region of the generated object. We then determine the generated object's circumscribed rectangle as its bounding box. The layout fidelity of the generated object with the ground-truth layout is quantified by the intersection-over-union (IoU) between the provided layout box and the generated object's circumscribed box. 

\section{Experiments}


\subsection{Model and Dataset} 

\noindent\textbf{Model} We leverage powerful pre-trained diffusion models as the foundation for our generative approach. Our best model is fine-tuned from Stable Diffusion XL (SDXL)~\cite{podell2023sdxl}. We also provide the benchmarks using Stable Diffusion 1.5 (SD1.5)~\cite{ldm}, which is a widely used backbone in existing methods~\cite{instdiff}. The proposed regional cross-attention layer is inserted into the original diffusion model right after each self-attention layer. The weights of the output linear layer are initialized to zero, ensuring that the model equals to the foundational model at the very beginning. More implementation details is shown in \Cref{appendix.implementation_details}

\noindent\textbf{Rich-Context Dataset} To equip the model with the capability to be conditioned on complex and detailed layout descriptions, a rich-context dataset is required. While obtaining large-scale real-world datasets through human tagging is labor-intensive and expensive, synthetic training data can be more readily acquired by leveraging recent advancements in large visual-language models. Similar to GLIGEN~\cite{li2023gligen} and InstDiff~\cite{instdiff}, we generate synthetic data to train our model. 

\begin{figure}[ht]
    \centering
    \begin{tabular}{@{}c@{}c@{}c@{}c@{}}
        \begin{tabular}[b]{c} \hspace{-0.4cm}
            \includegraphics[width=0.24\linewidth]{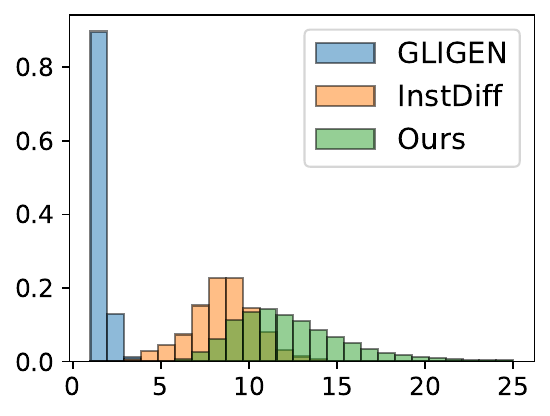} \\
            \small (a) Average Caption \\Length
        \end{tabular} & \hspace{-0.4cm}
        \begin{tabular}[b]{c}
            \includegraphics[width=0.24\linewidth]{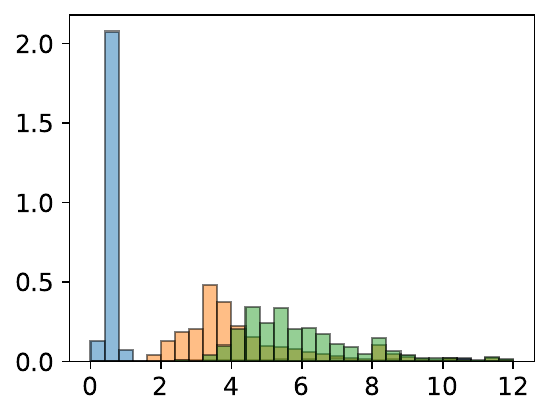} \\
            \small (b) Gunning Fog Score\\(Complexity)
        \end{tabular} & \hspace{-0.4cm}
        \begin{tabular}[b]{c}
            \includegraphics[width=0.24\linewidth]{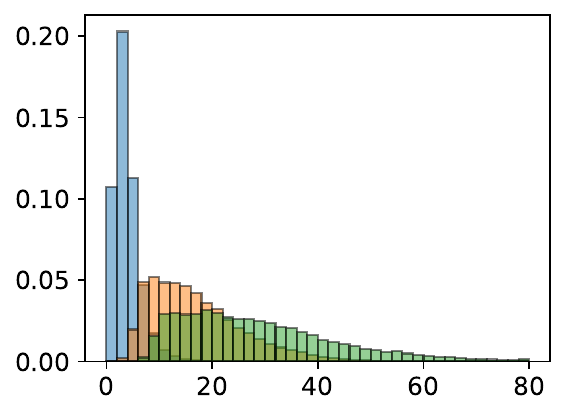} \\
            \small (c) Unique Words/Sample\\(Diversity)
        \end{tabular} & \hspace{-0.4cm}
        \begin{tabular}[b]{c}
            \includegraphics[width=0.24\linewidth]{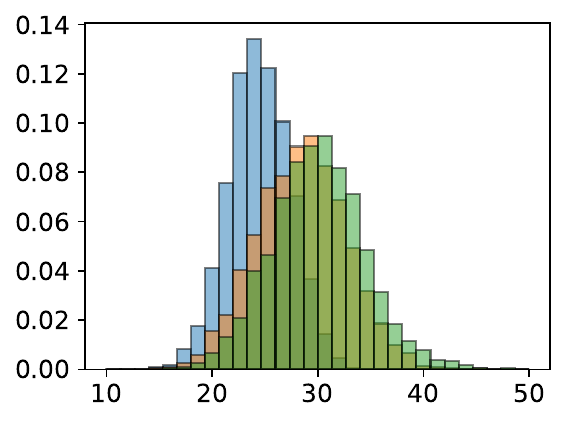} \\
            \small (d) Object-label CLIP\\Alignment Score
        \end{tabular}
    \end{tabular}
    \caption{Statistical comparisons between the synthetic object descriptions generated by GLIGEN~\cite{li2023gligen}, InstDiff~\cite{instdiff}, and our method. We measure the 1) average caption length, 2) the Gunning Fog Score, which estimates the text complexity from the education level required to understand the text, 3) the number of unique words per sample which indicates the text diversity, and 4) the object-label CLIP Alignment Score to measure object-label alignment. The results show that the pseudo-labels generated for our dataset are more complex, diverse, lengthier, and align better with objects, compared to those generated by GLIGEN and InstDiff.}
    \label{fig.statistic_dataset}
\end{figure}

We adopt a locating-and-labeling strategy during pseudo-label generation. At the first step, we use the Recognize Anything Model (RAM)~\cite{ram} and GroundingDINO~\cite{liu2023grounding} to identify and locate salient objects in the image. Next, we use the visual-language model QWen~\cite{qwen} to produce the synthetic label for each object by asking it to generate detailed description of the object (see \Cref{appendix.implementation_details} for the prompt we used). We utilize CC3M~\cite{cc3m} and COCO Stuff~\cite{coco} as the image source. For COCO, we directly use the ground-truth bounding boxes rather than relying on RAM and GroundingDINO to generate synthetic labels. The final training dataset contains two million images, with 10,000 images from CC3M set aside and the 5,000-image validation set of COCO used for evaluation. We denote the generated dataset Rich-Context CC3M (RC CC3M) and Rich-Context COCO (RC COCO). Compared to the synthetic training data used in GLIGEN~\cite{li2023gligen} and InstDiff~\cite{instdiff}, our rich-context dataset provides more diverse, complex, lengthy, and accurate descriptions, as shown in Figure~\ref{fig.statistic_dataset}. 

\begin{figure}[t]
    \centering
    \includegraphics[width=\linewidth,trim={0 0.3cm 0 0},clip]{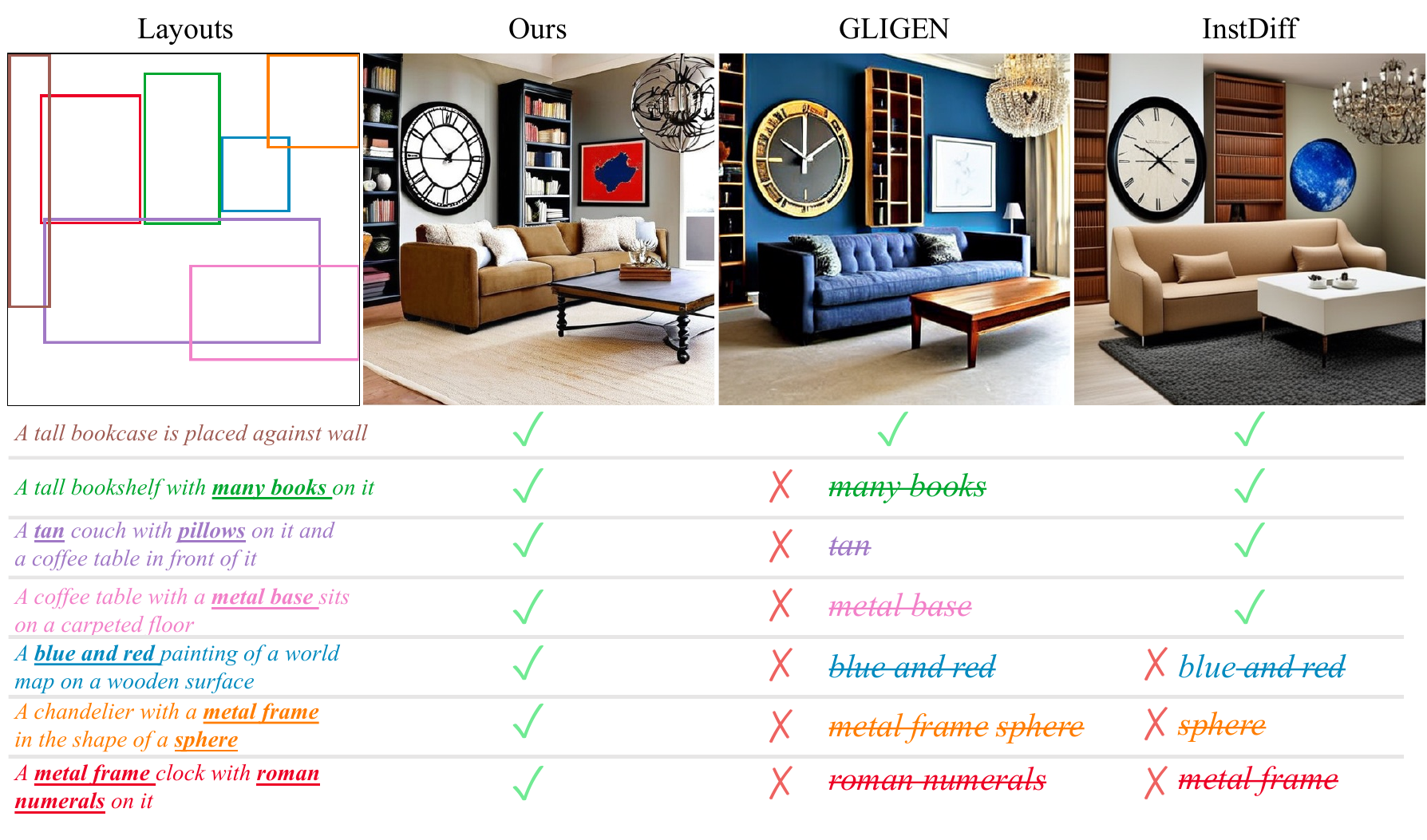}
    \caption{Qualitative comparison of rich-context L2I generation, showcasing our method alongside open-set L2I approaches GLIGEN~\cite{li2023gligen} and InstDiff~\cite{instdiff}, based on detailed object descriptions. Our method consistently generates more accurate representations of objects, particularly in terms of specific attributes such as colors and shapes. Strikethrough text indicates missing content in the generated objects from the descriptions. More qualitative results available in \Cref{appendix.more_results}}\label{fig.qualitative_comparison}
\end{figure}

\subsection{Reliability Analysis of Automatic L2I Evaluation}



Our proposed evaluation metrics in \Cref{sec.metric} serve as a substitute for the lack of precise ground-truth in open-set scenarios. Whether the set is closed or open, the goal of evaluation is to ensure that the measurement results align with human perception. To validate the reliability of our automatic evaluation metrics, we conducted a user study on the RC CC3M dataset, randomly selecting 1000 samples. Each synthetic sample may contain multiple objects, but only one object was randomly selected for each question. Users were asked to answer two questions, each rated on a scale from 0 (bad) to 5 (good). For object-label alignment, users responded to the question: "Can the cropped object in the image be recognized as [label]?" with the label being the automatically generated object description from RC CC3M. For layout fidelity, users answered: "How well (tight) does the object align with the bounding box?" referring to the synthetic bounding box in RC CC3M.

In total, we collected 300 answers for each question. We used the Pearson correlation coefficient to analyze how well the automatic evaluation metrics align with human perception. The Pearson correlation coefficient measures the correlation between two distributions with a value ranging from -1 to 1, where 0 indicates no relation and 1 indicates a strong correlation. Empirically, we found that the automatic evaluation metrics sometimes failed to reflect human perception when the object size was very small or very large. We note that for small objects, the clarity of the object can be hampered, while large objects may overlap with many other objects, making the automatic measurements inaccurate. Therefore, we filtered out objects smaller than 5\% or larger than 50\% of the image, resulting in an improved Pearson correlation between automatic metrics and user scores from 0.33 to 0.59 for CropCLIP and from 0.15 to 0.52 for SAMIoU. We applied the same filtering rule in the remaining evaluations. 


\subsection{Rich-Context Layout-to-Image Generation}

\noindent\textbf{Evaluation Metrics.} In addition to the two dedicated metrics discussed in \Cref{sec.metric}, we also consider image quality by measuring the FID scores~\cite{fid}, which reflects how real or natural the generated images look compared to real images. While we did not observe significant changes in FID scores across different variations of our methods, we did notice change in image quality among different baseline methods.

\noindent\textbf{Baseline Methods.} We compare our approach with GLIGEN~\cite{li2023gligen}, a popular open-vocabulary L2I generative model, and InstDiff~\cite{instdiff}, a recent method that achieves state-of-the-art open-set L2I performance. Besides, two training-free L2I methods BoxDiff~\cite{boxdiff} and R\&B~\cite{randb} are also considered for comparison. Although these methods can accept open-set words, their inputs are limited to single words or simple phrases which are not truly rich-context descriptions. Therefore, we denote them as constrained L2I methods and only evaluate them on COCO using category names.

\Cref{tab.quantitative} benchmarks the performance of L2I methods at an image sampling resolution of 512. Our model with SD1.5 achieves similar performance to InstDiff while reducing the computation cost in the layout conditioning layer by half, as illustrated in \Cref{fig.scatter}. Additionally, our model with SDXL achieves the best performance, even though the 512 resolution is sub-optimal for it. Further experiments in Section \Cref{sec.ablation} demonstrate that higher sampling resolutions can further enhance performance. \Cref{fig.qualitative_comparison} shows that, as the complexity and length of object descriptions increase, existing open-set L2I methods tend to overlook details  especially when objects are specified with colors or shapes. In contrast, our method consistently generates objects that accurately represent the given descriptions.

\begin{table}
  \caption{Quantitative comparison of different L2I approaches under image resolution at 512x512. `$\uparrow$' means that the higher the better, `$\downarrow$' means that the lower the better. }
  \label{tab.quantitative}
  \centering
  \small
  \begin{tabular}{lcccccc}
    \toprule
     & \multicolumn{3}{c}{\textbf{RC COCO}} & \multicolumn{3}{c}{\textbf{RC CC3M}}  \\
     \cmidrule(lr){2-4} \cmidrule(lr){5-7}
    \textbf{Method} & CropCLIP$\uparrow$ & SAMIoU$\uparrow$ & FID$\downarrow$  & CropCLIP$\uparrow$ & SAMIoU$\uparrow$ & FID$\downarrow$ \\
    \cmidrule(r){1-1} \cmidrule(lr){2-4} \cmidrule(lr){5-7}
    \multicolumn{1}{c}{\textit{Constrained}} \\
    BoxDiff~\cite{boxdiff} & 22.61 & 58.75 & 29.99 & - & - & -  \\ 
    R\&B~\cite{randb} & 23.68 & 64.68 & 31.70 & - & - & - \\ 
    \cmidrule(r){1-1} \cmidrule(lr){2-4} \cmidrule(lr){5-7}
    \multicolumn{1}{c}{\textit{Open-Set}} \\
    GLIGEN~\cite{li2023gligen} & 25.20 & 78.66 & 27.62 & 25.27 & 83.64 & \underline{15.81} \\ 
    InstDiff~\cite{instdiff} & \underline{27.46} & \underline{80.78} & 30.00 & \underline{28.46} & 85.59 & 17.96 \\ 
    \cmidrule(r){1-1} \cmidrule(lr){2-4} \cmidrule(lr){5-7}
    \multicolumn{1}{c}{\textbf{\textit{Ours}}} \\
    SD1.5 & 27.36 & \underline{80.78} & \textbf{25.81} & 28.45 & \underline{86.04} & 16.49 \\
    SDXL & \textbf{28.15} & \textbf{80.84} & \underline{27.41} & \textbf{29.42} & \textbf{86.56} & \textbf{11.02} \\
    \bottomrule
  \end{tabular}
\end{table}

\subsection{Performance Across Various Complexity of Object Descriptions}

By adopting pre-pooling textual features in the layout conditioning layer, our method maximizes the retention of textual information during generation. We observe that this design significantly enhances performance when dealing with complex and lengthy object descriptions. In \Cref{fig.scatter}(a), we categorize object description complexity using the Gunning Fog score into three levels: easy (scores 0-4), medium (5-8), and hard ($>$8). Additionally, we classify descriptions by length into phrases ($\leq$8 words), short sentences ($\leq$15 words), and long sentences ($\geq$16 words). Our results indicate that for simple and short descriptions, the performance difference between our method and state-of-the-art open-set L2I methods is close. However, as the complexity and length of the descriptions increase, our method consistently outperforms existing approaches.


\subsection{Ablation Study}\label{sec.ablation}

We investigate the effectiveness of the proposed region reorganization and the use of bounding box indicators on object-label alignment and layout fidelity using RC CC3M dataset. In experiments without region reorganization, we use straightforward averaging features for overlapping objects. Empirically, we observe that without region reorganization, our model struggles to generate the correct object when there is an overlap of objects with complex descriptions, leading to a significant drop in both object-label alignment and layout fidelity as shown in the \Cref{tab.ablation}.  


\begin{figure}
    \centering
    \begin{tabular}{@{}c@{}c@{}}
        \begin{tabular}[b]{@{}c@{}} \hspace{-0.4cm}
            \includegraphics[width=0.495\linewidth]{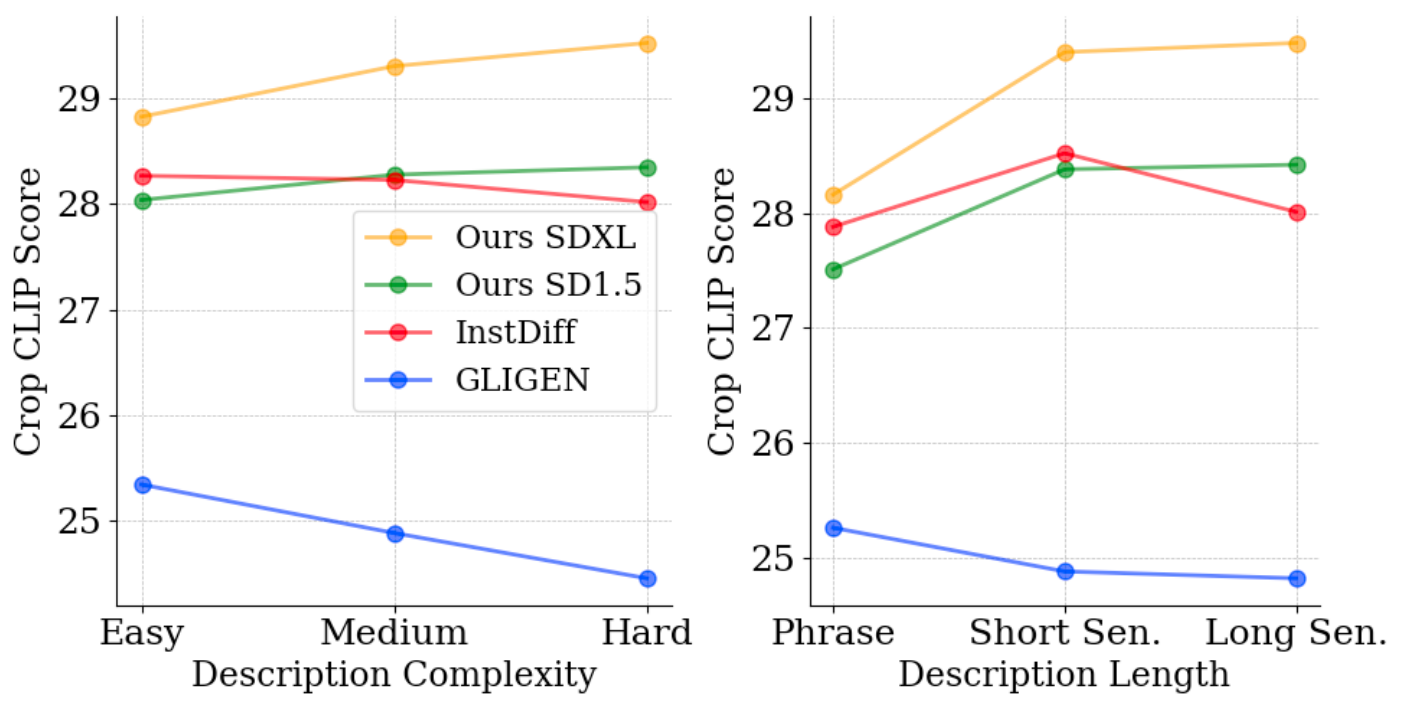} \\
            \small (a)
        \end{tabular} & \hspace{-0.4cm} 
        \begin{tabular}[b]{@{}c@{}}
            \includegraphics[width=0.495\linewidth]{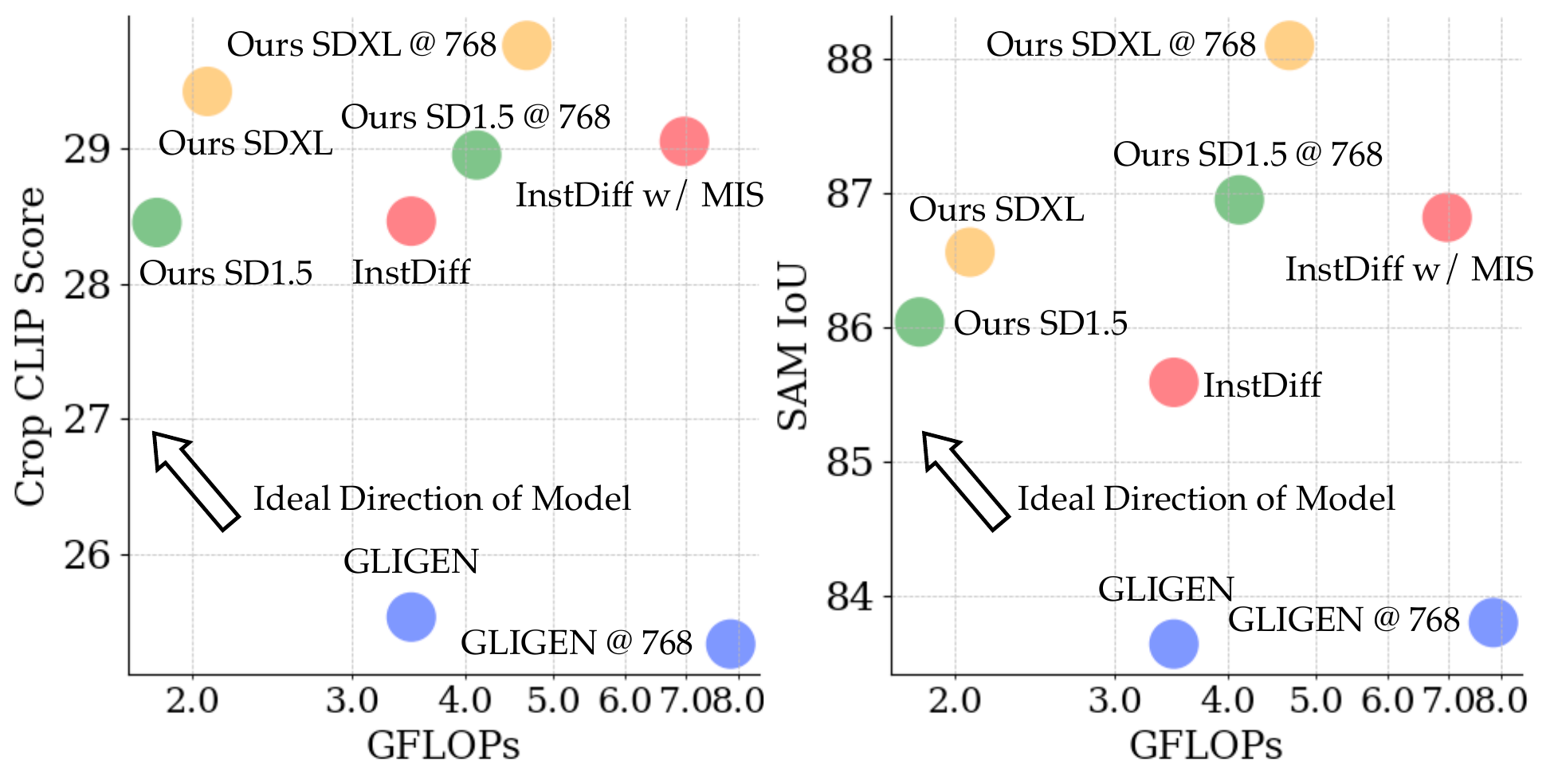} \\ 
            \small (b)
        \end{tabular}
    \end{tabular}
    \caption{(a) Object-text alignment scores across varying description complexities and lengths on RC CC3M. Our method shows significant advantages for complex and lengthy descriptions. (b) Object-text alignment and layout fidelity relative to computational cost in each layout-conditioning attention layer. Given that the number of textual tokens is much smaller than visual tokens, applying cross-attention can substantially reduce computational costs.}
    \label{fig.scatter}
    \vspace{-0.3cm}
\end{figure}

Unlike self-attention-based solutions that use box indicators to implicitly indicate object locations, our method explicitly cross-attends visual tokens with corresponding textual tokens. This approach allows the model to recognize the correct location for object conditioning even without a box indicator. However, the reorganized mask in the regional cross-attention layer has a lower resolution than the original image, causing misalignment near the borders of generated objects. Adding a bounding box indicator not only helps the model distinguish objects with similar descriptions and but also improves layout fidelity, as validated by the improvement in SAMIoU.

Additionally, we notice that sampling at a higher image resolution (768x768) improves model performance, although it demands greater computational resources. It's important to note that generalization to higher resolution is not a universal capability of L2I models. Existing self-attention-based L2I methods like GLIGEN~\cite{li2023gligen} experience performance declines when sampling at resolutions different from the training resolution. Another self-attention-based method, InstDiff~\cite{instdiff}, uses absolute coordinates for conditioning, requiring the sampling resolution to match the training resolution exactly. In \Cref{fig.scatter}(b), we compare the performance-computation trade-off\footnote{Computed using \url{https://github.com/MrYxJ/calculate-flops.pytorch} on an attention layer with 640 channels, which corresponds to a 32x32 resolution for image features, assuming input resolution of 512x512.} of open-set L2I approaches. Since InstDiff does not support flexible resolution sampling, we utilize Multi-Instance Sampling (MIS)~\cite{instdiff} instead. MIS was proposed to enhance InstDiff's performance by sampling each instance separately, albeit with increased inference times. We demonstrate the simplest case of MIS, which requires two inferences, but its computational cost scales linearly with the number of objects.


\begin{table}[!ht]
    \centering
    \small
    \caption{Ablation study of the proposed methods on RC CC3M dataset with SDXL backbone. The results suggest that region reorganization plays an important role for rich-context L2I generation, while using box indicator and sample at higher-resolution can further enhance performance.}
    \label{tab.ablation}
    \begin{tabular}{ccccc}
    \hline
    \textbf{Region Reorg.} & \textbf{Box Indicator} & \textbf{High Reso.} & CropCLIP$\uparrow$ & SAMIoU$\uparrow$ \\
    \cmidrule(r){1-1} \cmidrule(lr){2-2} \cmidrule(lr){3-3} \cmidrule(l){4-4} \cmidrule(l){5-5}
    \xmark & \xmark & \xmark & 25.32 & 76.92 \\
    \cmark & \xmark & \xmark & 28.93 & 85.02 \\
    \cmark & \cmark & \xmark & 29.42 & 86.56 \\
    \cmark & \cmark & \cmark & 29.79 & 88.10 \\
    \hline
  \end{tabular}
    
\end{table}


\section{Conclusion}

In this study, we introduced a novel approach to enhance layout-to-image generation by proposing Regional Cross-Attention module. This module improve the representation of layout regions, particularly in complex scenarios where existing methods struggle. Our method reorganizes object-region correspondence by treating overlapping regions as distinct standalone regions, allowing for more accurate and context-aware generation. Additionally, we addressed the gap in evaluating open-vocabulary L2I models by proposing two novel metrics to assess their performance in open-set scenarios. Our comprehensive user study validated the consistency of these metrics with human preferences. Overall, our approach improves the quality of generated images, offering precise location control and rich, detailed object descriptions, thus advancing the capabilities of generative models in various potential applications.

\textbf{Acknowledgement}
This work was funded in part by the Science and Technology Development Fund, Macau SAR (File no. 0049/2022/A1, 0050/2024/AGJ), and in part by the University of Macau (File no. MYRG2022-00072-FST, MYRG-GRG2024-00181-FST)

\bibliography{citation}
\bibliographystyle{ieee_citation_format}

\appendix

\newpage
\section*{\centering Supplementary Material for Rethinking The Training And Evaluation of Rich-Context Layout-to-Image Generation}

\section{Limitations}
Our work is built upon pre-trained diffusion models. Although our solution is backbone-agnostic, fine-tuning is still required when changing the model's backbone. Additionally, our training dataset is generated using visual-language model, which cannot guarantee that all synthetic labels are correct; these inaccuracies may negatively impact the model's performance. Furthermore, our training images are from publicly available datasets, which often contain low-quality images. As a result, the generated images may exhibit undesired artifacts, such as watermarks, during generation.

\section{Implementation Details}~\label{appendix.implementation_details}
We train our model using the AdamW~\cite{adamw} optimizer with a learning rate of 5e-5. The training process involves an accumulated batch size of 256, with each GPU handling a batch size of 2 over 8 accumulated steps, for a total of 100,000 iterations on 16 NVIDIA V100 GPUs. This training process takes approximately 8,000 GPU hours. During training, we apply random cropping and horizontal flip for image augmentation, a bounding box will be dropped if its remaining size is smaller than 30\% of its original size after cropping. We randomly drop 10\% of layout conditions (all conditions in an image are dropped when a layout condition is dropped) and 10\% of image captions to support classifier-free guidance~\cite{cfg}. During sampling, we use a classifier-free guidance scale of 4.5 for our SDXL-based model and 7.5 for our SD1.5-based model. The inference denoising step is set to 25 for our models and all baseline methods. During synthetic data generation, we obtain the description of the object using the following prompt for QWen model: ``You are viewing an image. Please describe the content of the image in one sentence, focusing specifically on the spatial relationships between objects. Include detailed observations about all the objects and how they are positioned in relation to other objects in the image. Your response should be limited to this description, without any additional information''.

\section{Throughput of Different Layout-to-Image Methods. }

\begin{figure}[!h]
    \centering
    \includegraphics[width=0.5\linewidth]{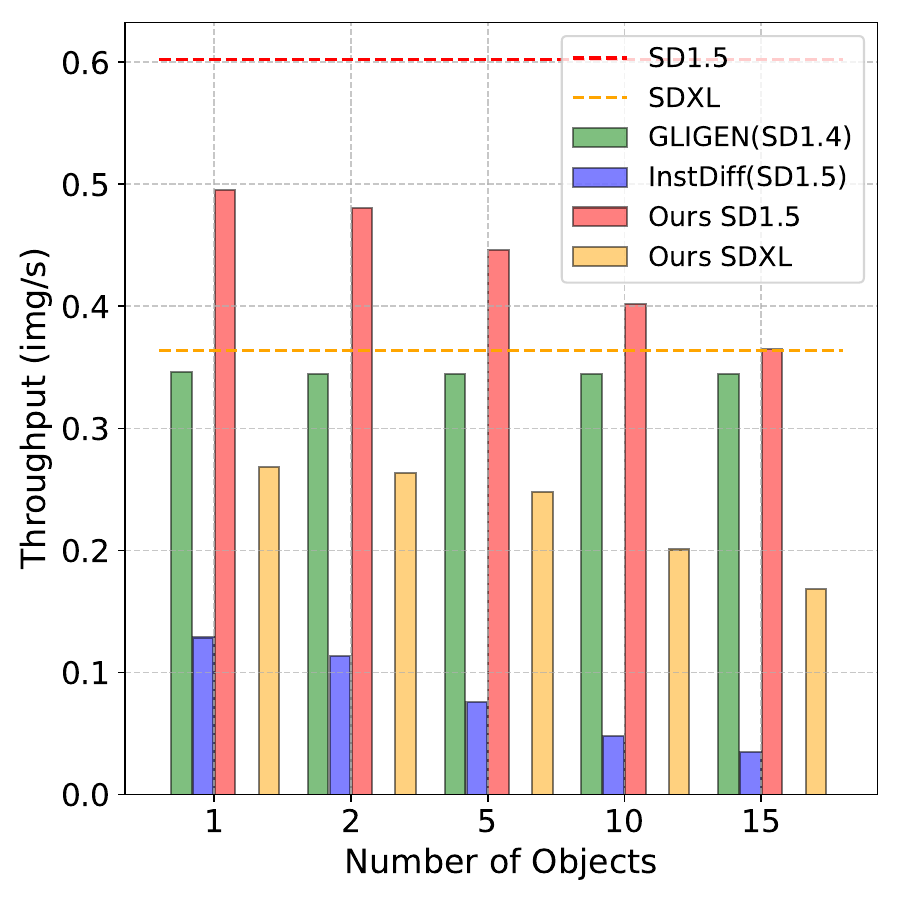}
    \caption{All methods are tested with float16 precision and 25 inference steps. The results are averaged over 20 runs. Notably, the overall throughput of our method is not significantly hampered. In a typical scenario with 5 objects, the throughput of our method exceeds 60\% of the throughput of the original backbone model. Please note that while the official backbone of GLIGEN is SD1.4, its network structure and throughput are identical to those of SD1.5.}
    \label{fig:throughput}
\end{figure}

In addition to the FLOPs comparison presented in \Cref{sec.ablation}, we compare the throughput of using different L2I methods and present the result in the \Cref{fig:throughput}

\section{Layout-to-Image Generation Diversity Comparison}
Following LayoutDiffusion~\cite{zheng2023layoutdiffusion}, we evaluate the generation diversity using LPIPS and Inception Score and present the diversity comparison of different L2I methods  using 1,000 RC CC3M evaluation layouts in \Cref{tab:comparison}. 

\begin{table}[h!]
\centering
\begin{tabular}{lcc}
\toprule
\textbf{} & \textbf{LPIPS}$\uparrow$ & \textbf{Inception Score}$\uparrow$ \\
\midrule
GLIGEN & 0.65 $\pm$ 0.10 & 15.74 $\pm$ 1.52 \\
InstDiff & 0.67 $\pm$ 0.10 & 15.26 $\pm$ 1.75 \\
Ours (SD1.5) & \textbf{0.71} $\pm$ 0.13 & \textbf{16.95} $\pm$ 2.31 \\
Ours (SDXL) & \underline{0.68} $\pm$ 0.12 & \underline{15.86} $\pm$ 1.58 \\
\bottomrule
\end{tabular}
\caption{For LPIPS computation, each layout is inferred twice, and the score is calculated using AlexNet. A higher LPIPS score indicates a larger feature distance between two generated images with the same layouts, signifying greater sample-wise generation diversity. A higher Inception Score suggests a more varied appearance of generated images, indicating greater overall generation diversity.}
\label{tab:comparison}
\end{table}

\section{Pseudo-code For Proposed Evaluation Metrics}

\Cref{alg:cropclip,alg:samiou} shows the pseudo-code for computing the proposed Crop CLIP score and SAMIoU score on a single generated sample. The final performance is calculated by averaging these scores across all generated samples.

\begin{algorithm}
\caption{Compute Crop CLIP Score}\label{alg:cropclip}
\begin{algorithmic}[1]
\Require Generated image \texttt{I}, conditioning layout boxes \texttt{B} and labels \texttt{L} for each object, CLIP models $\mathrm{clip}_{img}$ and $\mathrm{clip}_{text}$

\State \texttt{crop\_clip\_scores} $\gets$ \texttt{[]}
\For{each $(\texttt{box}, \texttt{label}) \in \text{zip}(\texttt{B}, \texttt{L})$}
    \State \texttt{S} $\gets$ $\mathrm{crop}$(\texttt{I}, \texttt{box})
    \If {\texttt{S} size $< \text{lower thres.}$ \textbf{or} \texttt{S} size $> \text{upper thres.}$}
        \State \textbf{continue}
    \EndIf
    \State \texttt{clip\_img\_feat} $\gets$ $\mathrm{clip}_{img}$(\texttt{S})
    \State \texttt{clip\_text\_feat} $\gets$ $\mathrm{clip}_{text}$(\texttt{label})
    \State \texttt{crop\_clip\_sim} $\gets$ $\mathrm{cosine\_similarity}$(\texttt{clip\_img\_feat}, \texttt{clip\_text\_feat})
    \State \texttt{crop\_clip\_scores}.append(\texttt{crop\_clip\_sim})
\EndFor
\State \texttt{sample\_crop\_clip\_score} $\gets$ $\mathrm{mean}$(\texttt{crop\_clip\_scores})
\State \textbf{return} \texttt{sample\_crop\_clip\_score}
\end{algorithmic}
\end{algorithm}

\begin{algorithm}
\caption{Compute SAMIoU Score}\label{alg:samiou}
\begin{algorithmic}[1]
\Require Generated image \texttt{I}, conditioning layout boxes \texttt{B} for each object, Segment Anything Model $\mathrm{SAM}$

\State \texttt{sam\_iou\_scores} $\gets$ []
\For{$\texttt{box} \in B$}
    \If {\texttt{S} size $< \text{lower thres.}$ \textbf{or} \texttt{S} size $> \text{upper thres.}$}
        \State \textbf{continue}
    \EndIf
    \State \texttt{sam\_mask} $\gets$ $\mathrm{SAM}$(\texttt{I}, \texttt{b})
    \State \texttt{box\_of\_generated\_obj} $\gets$ $\mathrm{get\_circumscribed\_rectangle}$(\texttt{sam\_mask})
    \State \texttt{sam\_iou} $\gets$ $\mathrm{compute\_IoU}$(\texttt{box}, \texttt{box\_of\_generated\_obj})
    \State \texttt{sam\_iou\_scores}.append(\texttt{sam\_iou})
\EndFor
\State \texttt{sample\_sam\_iou\_score} $\gets$ $\mathrm{mean}$(\texttt{sam\_iou\_scores})
\State \textbf{return} \texttt{sample\_sam\_iou\_score}
\end{algorithmic}
\end{algorithm}

\section{Effectiveness of Rich-Context Dataset and Regional Cross-Attention}

The ablation study in \Cref{tab:ablation_rich_context_and_module} indicates that to condition the L2I model with rich-context descriptions, both a rich-context dataset and a designated conditioning module for rich-context description are vital.

\begin{table}[h!]
\centering
\small
\begin{tabular}{cccccccc}
\toprule
\multirow{3}{*}{\textbf{Backbone}} & \multicolumn{2}{c}{\textbf{Dataset}} & \multicolumn{3}{c}{\textbf{Attention Module}}  &  \multirow{3}{*}{\textbf{CropCLIP}} & \multirow{3}{*}{\textbf{\textbf{SAMIoU}}} \\ 
 \cmidrule(lr){2-3} \cmidrule(lr){4-6} 
& \multirow{2}{*}{Word/Phrase}  & \multirow{2}{*}{Rich-context} & SelfAttn & SelfAttn & CrossAttn  \\ 
& & & GLIGEN & InstDiff & Ours & & \\
\cmidrule(r){1-1} \cmidrule(lr){2-3} \cmidrule(lr){4-6} \cmidrule(lr){7-7} \cmidrule(l){8-8} 

SDXL & \checkmark & &  & & \checkmark & 25.40 & 86.76 \\ 
SDXL & & \checkmark & & & \checkmark & 29.79 & 88.10 \\ 

\cmidrule(r){1-1} \cmidrule(lr){2-3} \cmidrule(lr){4-6} \cmidrule(lr){7-7} \cmidrule(l){8-8} 

SD1.5 & & \checkmark & \checkmark & &  & 25.56 & 82.72 \\ 
SD1.5 & & \checkmark & & \checkmark &  & 28.36 & 85.58 \\ 
SD1.5 & & \checkmark & & & \checkmark & 28.94 & 86.91 \\ 
\bottomrule
\end{tabular}
\caption{The performance is evaluated on RC CC3M evaluation set and all methods are sampled under their best sampling resolution as discussed in \Cref{sec.ablation}. It can be noticed that even with the rich-context dataset, the performance of self-attention-based modules does not show significant improvement over their performance in the \Cref{tab.quantitative} in the paper.}
\label{tab:ablation_rich_context_and_module}
\end{table}

\section{Effectiveness of Regional Cross-Attention with Visual Example}

\begin{figure}[ht]
    \centering
    \includegraphics[width=\linewidth]{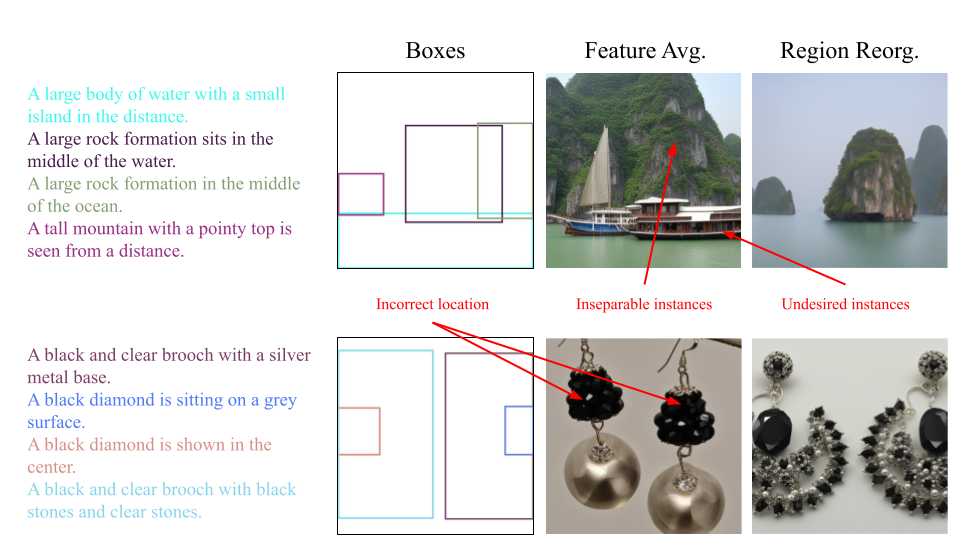}
    \caption{The model with region reorganization can accurate generated objects that better align with the designated layouts, while the feature averaging solution can result in objects in incorrect location, generating undesired instances or making the overlapping instances inseparable.}
    \label{fig:vs_avg}
\end{figure}

\Cref{fig:vs_avg} presents the visual comparison of using region reorganization compared to straightforward feature averaging when dealing with overlapping objects. The model with region reorganization can accurate generated objects that better align with the designated layouts, while the feature averaging solution can result in objects in incorrect location, generating undesired instances or making the overlapping instances inseparable.

\section{More Qualitative Results}~\label{appendix.more_results}

\begin{figure}[t]
    \centering
    \includegraphics[width=\linewidth]{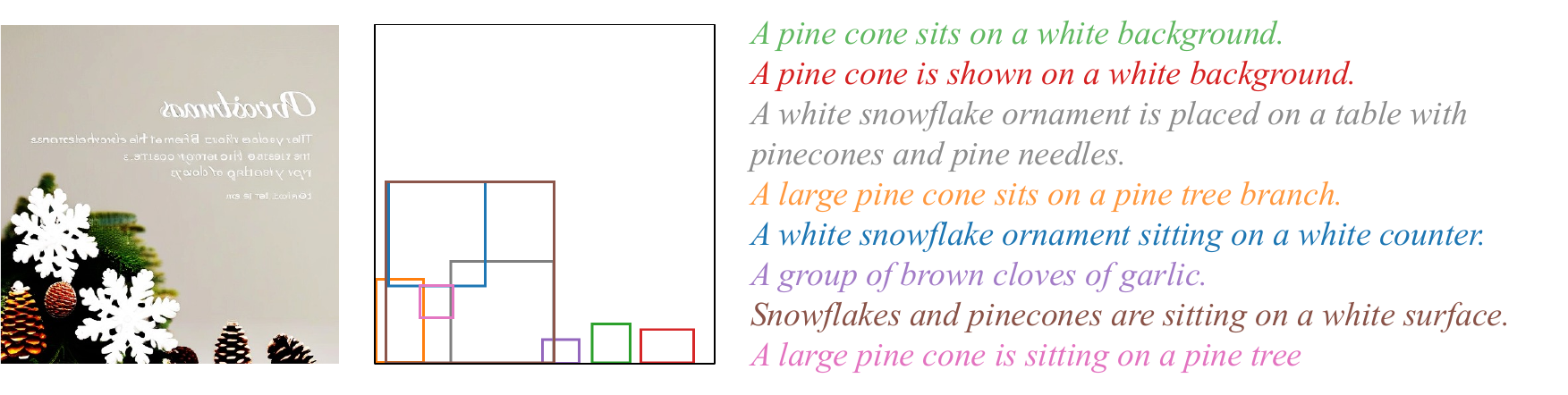}
    \includegraphics[width=\linewidth]{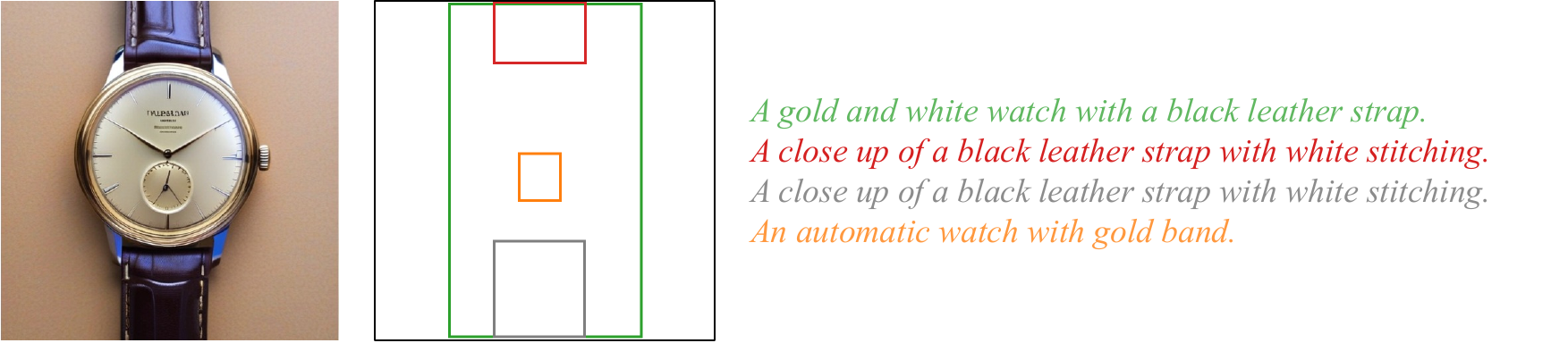}
    \includegraphics[width=\linewidth]{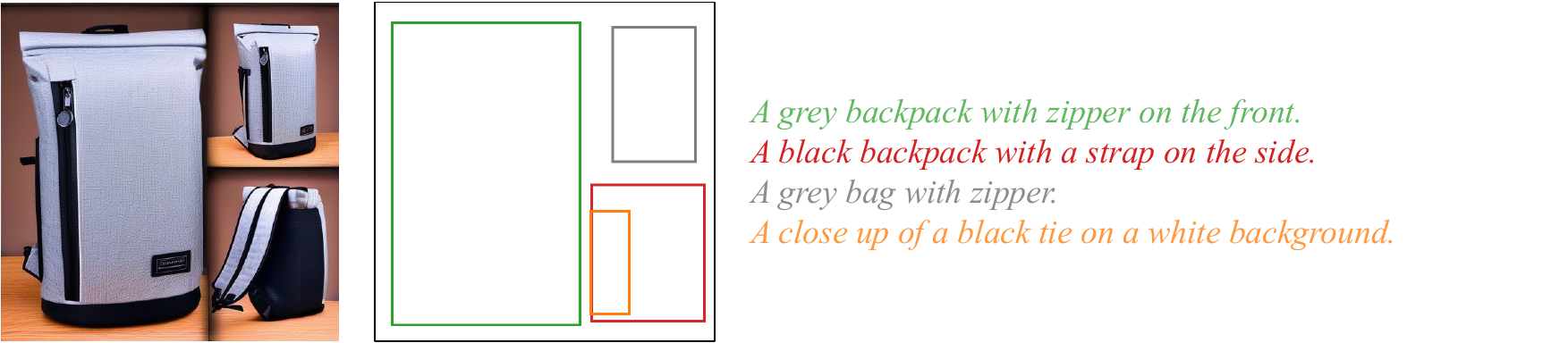}
    \includegraphics[width=\linewidth]{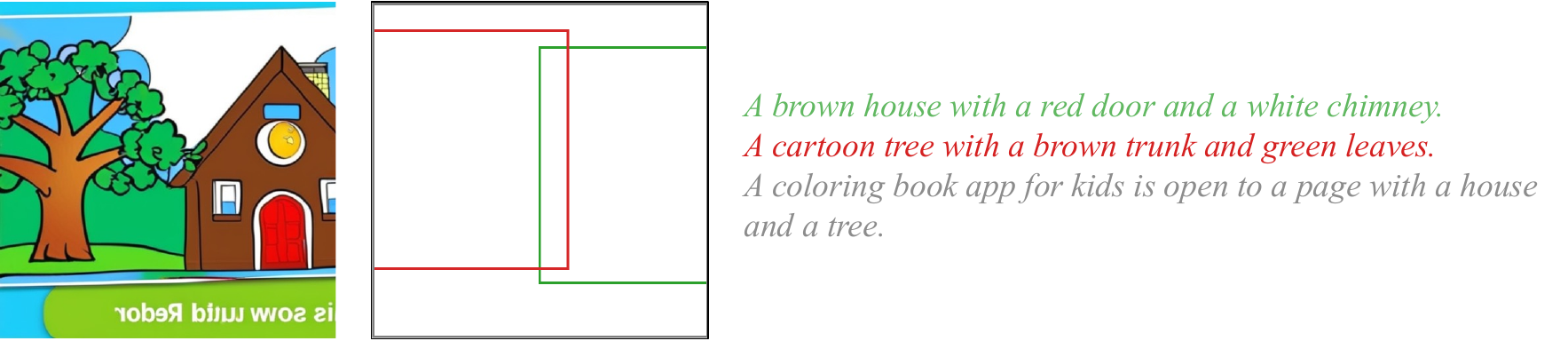}
    \includegraphics[width=\linewidth]{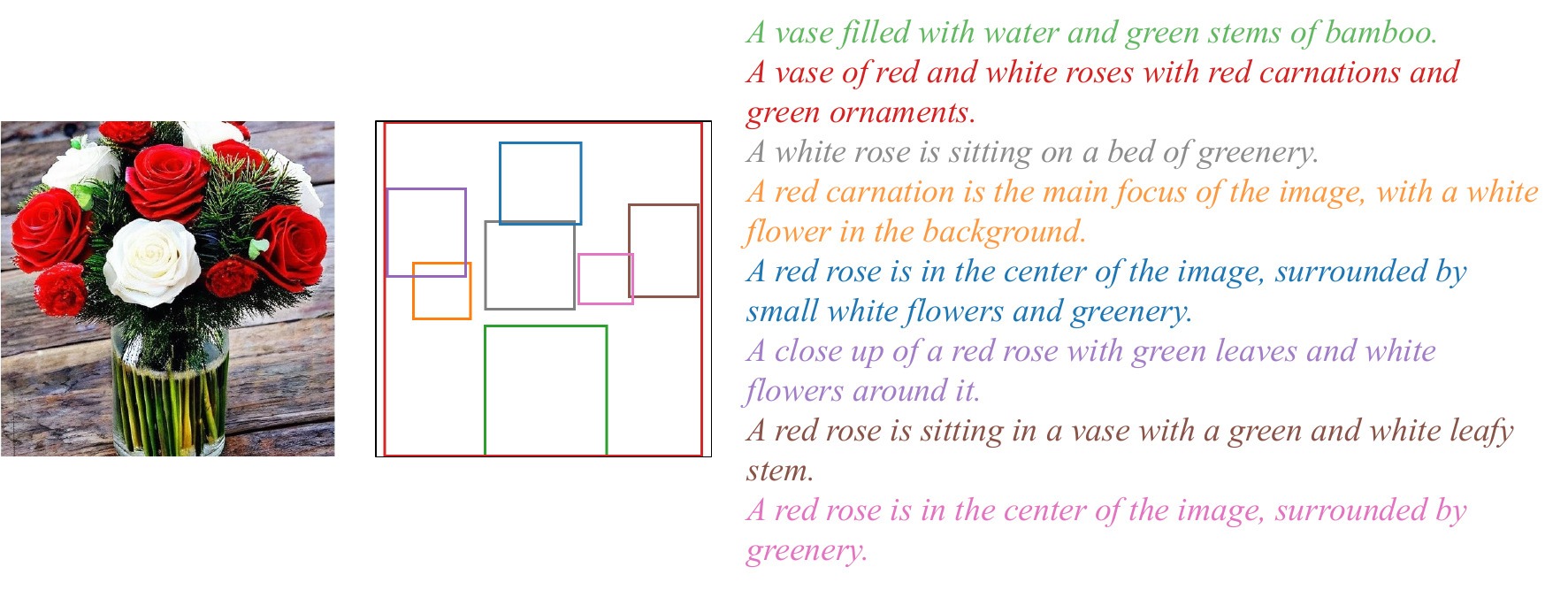}
    \caption{Additional qualitative results using random layouts from the synthetic RC CC3M dataset demonstrate our model's ability to accurately handle both very large and very small objects.}\label{fig.additional_viual}
\end{figure}

During the evaluation, we filtered out very small and very large regions to avoid inconsistencies between automatic evaluations and human preferences. However, this does not imply that our method is incapable of generating high-quality small or large objects. Our results in \Cref{fig.additional_viual} verify that our model can accurately handle both very large and very small objects.

\begin{figure}[t]
    \centering
    \includegraphics[width=\linewidth]{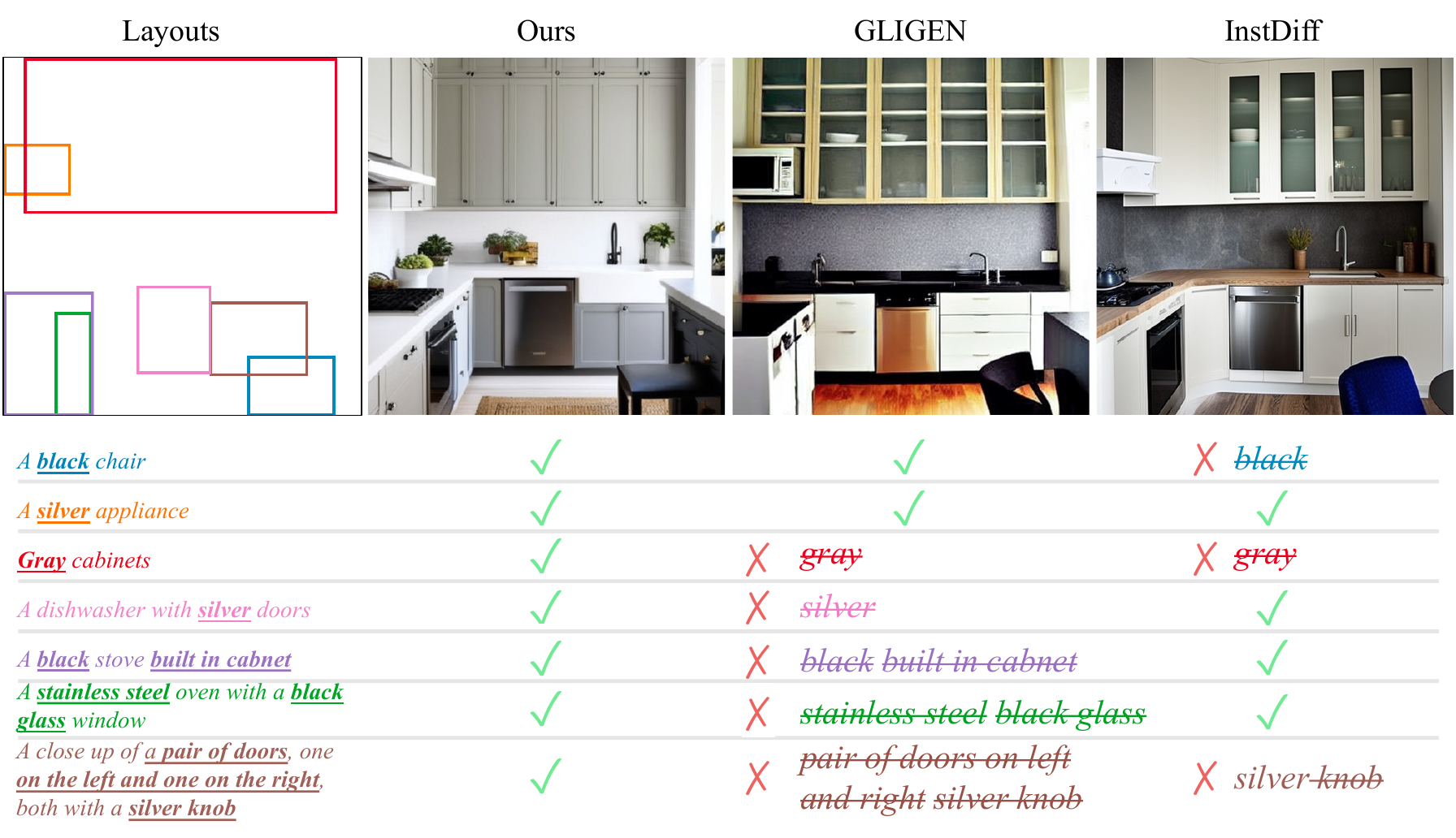} 
    \includegraphics[width=\linewidth]{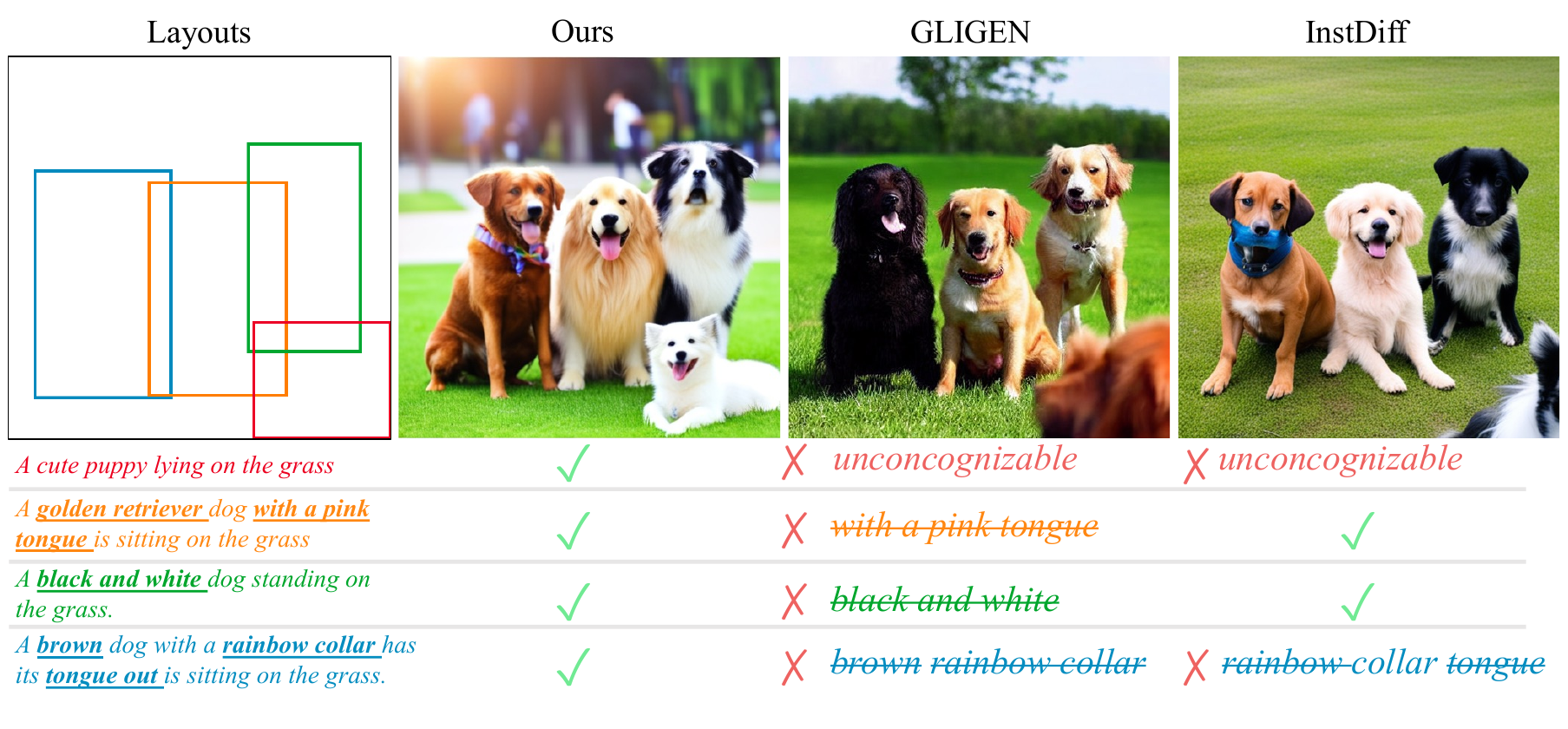} 
    \caption{More qualitative comparison of rich-context L2I Generation, showcasing our method alongside open-set L2I approaches GLIGEN~\cite{li2023gligen} and InstDiff~\cite{instdiff}, based on detailed object descriptions. Our method consistently generates more accurate representations of objects, particularly in terms of specific attributes such as colors and shapes. Strikethrough text indicates missing content in the generated objects from the descriptions.}\label{fig.more_qualitative_comparison}
\end{figure}

In addition, We provide more qualitative comparison with existing open-set L2I approaches in \Cref{fig.more_qualitative_comparison}. 

\end{document}